# A Robotic Medical Clown (RMC): Forming a Design Space Model

Liberman-Pincu E, Oron-Gilad T.
elapin@post.bgu.ac.il; orontal@bgu.ac.il
Human-Robot Interaction Laboratory
Dept. of Industrial Engineering and Management
Ben-Gurion University of the Negev
Beer-Sheva 84105, Israel

**Medical clowns help hospitalized children in reducing pain and anxiety symptoms and increase the level of satisfaction in children's wards. Unfortunately, there is a shortage of medical clowns around the world. Furthermore, isolated children can not enjoy this service. This study explored the concept of a Robotic Medical Clown (RMC) and its role. We used mixed methods of elicitation to create a design space model for future robotic medical clowns. We investigated the needs, perceptions, and preferences of children and teenagers using four methods: interviewing medical clowns to learn how they perceive their role and the potential role of an RMC, conducting focus groups with teenagers, a one-on-one experience of children with a robot, and an online questionnaire. The concept of RMCs was acceptable to children, teenagers, and medical clowns. We found that the RMC's appearance affects the perception of its characters and role. Future work should investigate the interaction in hospitals.**

## I. INTRODUCTION

Medical clowning was found to be effective in reducing pain and anxiety symptoms among hospitalized children [1-3]. In addition, it increases the level of satisfaction with treatment in children's wards [4]. This service cannot be provided to isolated children; furthermore, there is a shortage of medical clowns around the world. Hence, new solutions must be found. Socially assistive robots (SARs) have been found to be acceptable among children [5] and beneficial in various domains such as education, companionship, or physical exercise [6-9]; therefore, Robotic Medical Clowns (RMCs) may provide emotional and communicative support, addressing some of these issues and complementing human medical clowns.

Robots' appearance and performance should be carefully designed to gain children's engagement and build a positive child-robot relationship [10]. Our previous studies assessed the effect of robots' visual qualities (e.g., body structure, color, etc.) on user perceptions and revealed the importance of matching a robot's appearance to its context of use [11,12]. Even minor design manipulations affected users' perceptions and behaviors [13].

In this study, we sought to create a design space model for future robotic medical clowns. We investigated the needs, perceptions, and preferences of children and teenagers regarding three design factors of the RMCs: functionality, behavior, and appearance.

## II. METHODOLOGY

To gain a broad perspective of the role and potential use of a robotic medical clown, our analysis includes different stakeholders. We used a mixed-methods design; interviewing medical clowns, focus groups with teenagers, a one-on-one experience of children with an RMC prototype, and an online questionnaire. The outcomes of each step provide the foundations for the next one. Figure 1 illustrates the study design and workflow. The following sections detail each step.

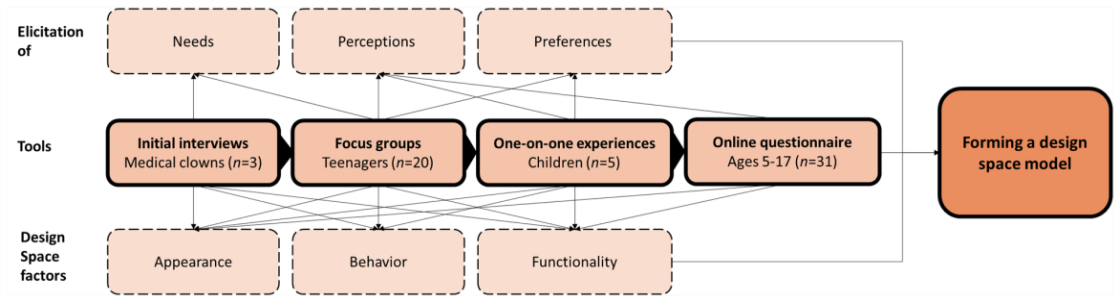

Figure 1: Study design and workflow.

*1: Initial interviews*

Our first step in this research was to conduct preliminary interviews with certified medical clowns of the "Simchat HaLev" association who volunteer in pediatric wards in hospitals in Israel. The interviews aimed to understand the relationships between the child (patient), the caregiver, and the medical clown. And to learn more about different possible activities in the hospital setting. The interviews were structured following the three design factors, functionality, behavior, and appearance, and were parsed into five themes: the primary purpose of medical clowning, activities and interactions with a medical clown, what they thought would be suitable activities and interactions for an RMC, and the appearance of medical clowns and RMCs.

*2: Focus groups*

To gain the perspectives of teenagers on what they think can be activities and interactions for an RMC, we conducted two structured focus groups with a convenience sample of middle and high school students from two cities in Israel, members of robotic-enthusiastic teams. First, we presented the *temi robot* (One Robotics) on which the robotic medical clown prototype would be applied. Then, we explained the goals of the meeting. The discussion was divided into three parts: what they thought hospitalized children's feelings and emotions were, options for games and activities children can play with/on the RMC for the hospital settings and the appropriate appearance for an RMC.

*3: One-on-one experiences*

In this part, we sought to investigate the child-robot interaction in a lab setting. We implemented the insights from the interviews and focus groups on the *temi robot* and designed an RMC prototype using the Android Studio platform. The experience included a short conversation with the RMC, selecting a game to be played on the RMC's screen, and an active engagement using the RMC's ability to move around and follow the child. After the experience, the participants were asked to fill out a printed notebook questionnaire to express their perceptions and preferences. The notebook aimed to collect insight into four themes: *Overall experience, Appearance perception and preferences, RMC's desired character,* and *Imaginary robots.*

*4: Online questionnaire*

The online questionnaire aimed to evaluate the RMC's appearance's effect on children's perception of its character and role. Based on the previous stages, we created four different appearances for the RMC: original appearance, clown-like, dog-like, and child-like (Figure 2). We used an online questionnaire. Respondents were exposed to only one of four possible design manipulations. A snowball distribution strategy was applied via social media and WhatsApp.

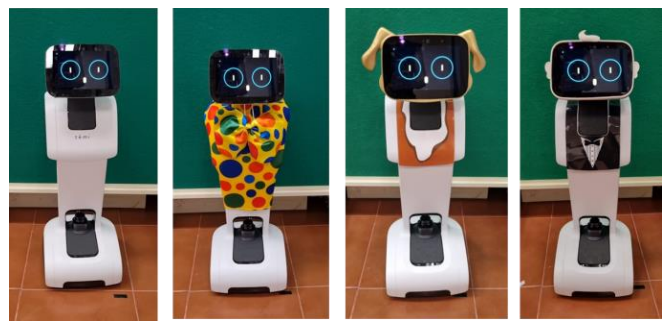

Figure 2: Four different appearances for the RMC: original appearance, clown-like, dog-like, and child-like.

Following a short video presenting the RMC and its possible activities with a child (talk, play); click here to watch the video), respondents were asked to fill out a two parts questionnaire: (A) Perceived characteristics and role and (B) Acceptance (based on [14]).

## III. RESULTS

*1: Initial interviews*

Three interviews were conducted with certified medical clowns (two men and one woman). The interviews took place in April 2022. Each interviewee individually for about 50 minutes. The interviews were recorded, transcribed, and analyzed using thematic analysis. Table 1 summarizes the main findings.

Table 1: Main findings from the interviews with medical clowns

| Theme | Main findings |
|---|---|
| **Functionality** Medical clowning's purpose | Comforting the patient and the caregiver Distraction Reducing pain through laughter |
| **Behavior Medical Clown** Current activities and interactions | Addressing the patient by name Playing games Using magic Referring also to the caregiver/guardian of the child |
| **Behavior RMC** Future possible activities and interactions with an RMC | Creating a dialogue Using humor The patient should be active in the interaction (e.g., by selecting games) The patient should be active physically in the room The RMC may leave a small gift |
| Medical clowns' **appearance** | Medical clowns wear clown suits, makeup, and a red nose. |
| RMC's **appearance** | It should appeal to children; consider using clown-like, human-like, or pet-like figures. |

*2: Focus groups*

We conducted two focus groups, each with a mixed group of 10 teenagers (ages 13-18) from the two teams who attended a seminar on robotics at Ben Gurion University. First, the participants talked about hospitalized children's feelings, fears, and desires and discussed the concept of using robots as medical clowns. Following this, they suggested optional physical and cognitive activities with the robots.

Regarding the RMC's appearance, both groups agreed it should be human-resembled and suggested using human clothes. The outcome of these focus groups led to the design of the one-on-one experience study and the RMC temi robot prototype. Figure 3 presents one of the sessions and illustrates some of the insights.

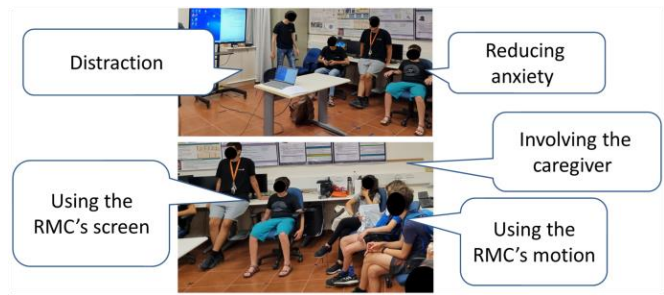

Figure 3: Illustration of the participant of the focus group suggesting activities

*3: One-on-one experiences*

Five participants (children aged 8-12 accompanied by their parents) were recruited voluntarily for a one-on-one experience with the temi robot in the lab. The overall experience with the robotic medical clown experience lasted about 20-30 minutes.

During the session, the participant was asked to try three different applications of the RMC prototype. First, the interaction started with a short conversation; the RMC introduced himself and asked for the child's name. Then, the RMC asked the participants to select games they would like to play using the RMC's screen out of a list of different options, including puzzles, drawings, and action. The final phase was an active engagement with the child and the parent/guardian, playing hide and seek. Figure 4 presents the three different activities the participant tried.

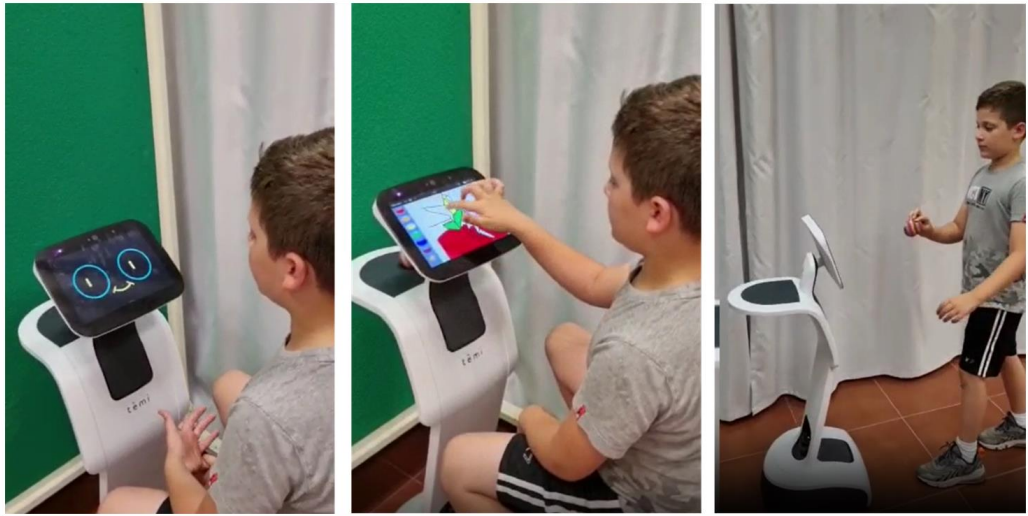

Figure 4: Left to right: the child is interacting with the RMC (greetings), the child is playing a drawing game, and the child is playing hide and seek.

Following the experience, the participants filled out a printed notebook questionnaire addressing four themes: *Overall experience, Appearance perception and preferences, RMC's desired character,* and *Imaginary robots.* As detailed below.

***Overall experience.*** We used a user experience questionnaire measuring enjoyment, intention to use, and attitude toward technology. Participants rated their overall experience as very positive; perceived enjoyment and intention-to-use were rated 4.8 out of 5, on average. Attitude toward technology was rated 4.6 on average. Figure 5 presents one example of the user experience questionnaire.

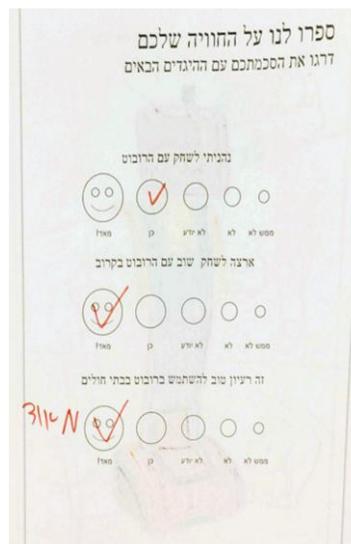

Figure 5: the user experience page of the notebook questionnaire

***Appearance perception and preferences***. We used a drawing of the robot as a coloring page where participants were asked to paint and design as they desired. All the children chose to use many colors. They expressed their desire for human-like figures, drew a human face onto the robot's screen, added a clown suit, and one girl even designed the robot as a female robot. Figure 6 presents two examples drawn by two of the participants.

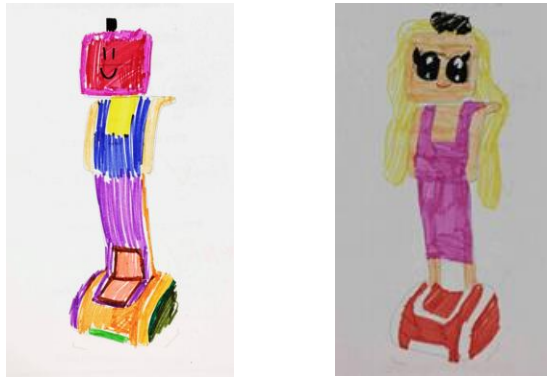

Figure 6: Two examples of the coloring sheet of the temi robot drawn by two participants.

*RMC's desired character.* We used blank comics presenting an interaction between an RMC and a hospitalized child; the participants were asked to add text to their conversation. This tool was used to collect insights regarding the children's expectations of RMCs and the desired characteristics. Analyzing their dialogs, we found that they were seeking comforting words and compassion (the robot is approaching the child and asking how he feels, wishing the child to feel better soon). They wanted this, in addition to more practical functions such as offering to play a game or tell a joke (either the robot offers or the child asks for it) or even to bring the child a cup of tea. Figure 7 presents one example of a comic drawn by a participant.

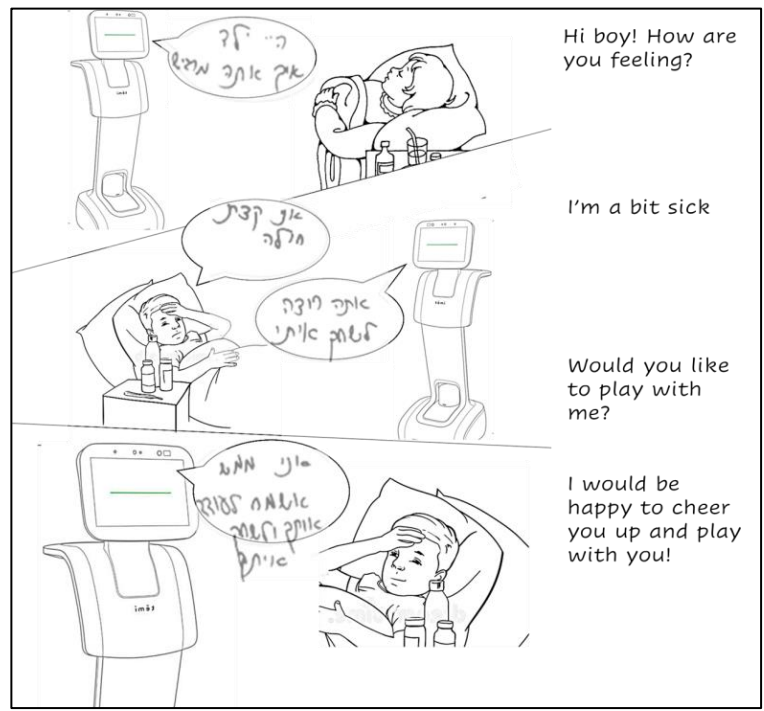

Figure 7: one example of a comic drawn by a participant. The translated text appears on the right side.

*Imaginary robots.* We asked the children to invent a new imaginary robot by text and drawings. This part aimed to understand children's perceptions and preferences of robots' roles and functionalities. The participants chose different morphologies for their designs: human-like, animal-like, and even a heart shape robot with a human face. Figure 8 presents one example of an imaginary robot—the lion robot.

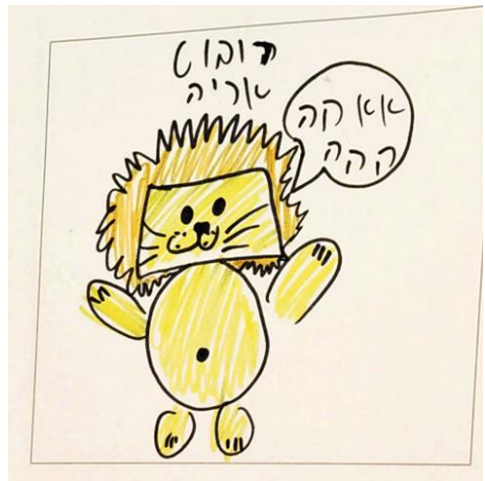

Figure 8: The lion robot- an example of an imaginary robot drawn by one of the participants.

*4: Online questionnaire*

Thirty-three respondents (aged 5-17) completed the online questionnaire. Results indicated that the appearance of the RMC moderately affected the children's perception of its character, the participants' acceptance, and the perception of its role. The clown-like RMC was rated as the least innovative, elegant, friendly, and authoritative. The child-like RMC was perceived as the most authoritative and innovative. The dog-like RMC was perceived as slightly more threatening than the other three designs. Figure 9 presents the perceived characteristics by the appearances.

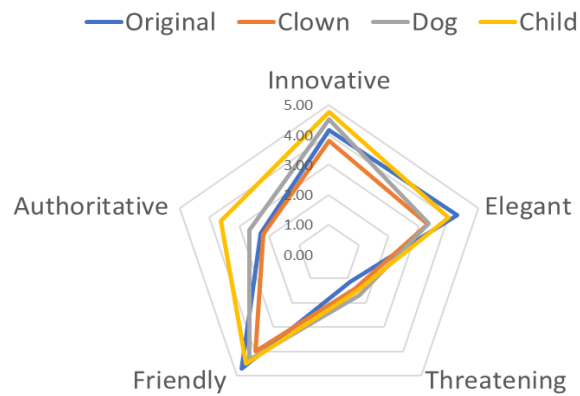

Figure 2: Perceived characteristics for RMCs [1-5] from the online questionnaire.

The child-like RMC was rated higher than the other three appearances in three out of the four categories of acceptance: attitude towards technology, perceived enjoyment, and perceived usefulness. It was rated the lowest in the fourth category, perceived ease of use.; the dog-like category was ranked highest in this category. The clown-like RMC was rated lowest in the perceived usefulness category. Figure 10 presents the measured acceptance of the different RMCs' appearances.

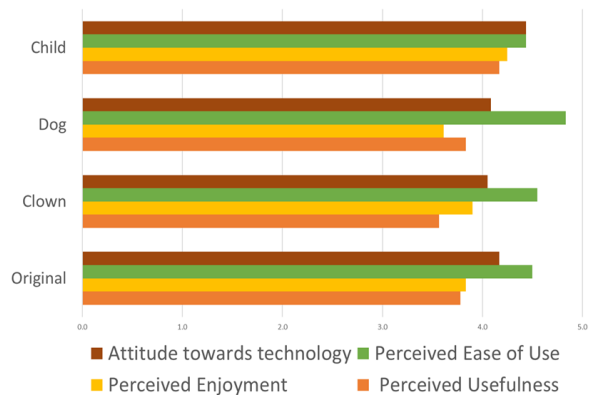

Figure 3: Measuring acceptance [1-5] from the online questionnaire.

The child-like robot was perceived as the one who could fulfill different roles. On the opposite, the clown-like robot was perceived as the least suitable for different tasks. Figure 11 presents the children's perception of the RMC's role by appearance.

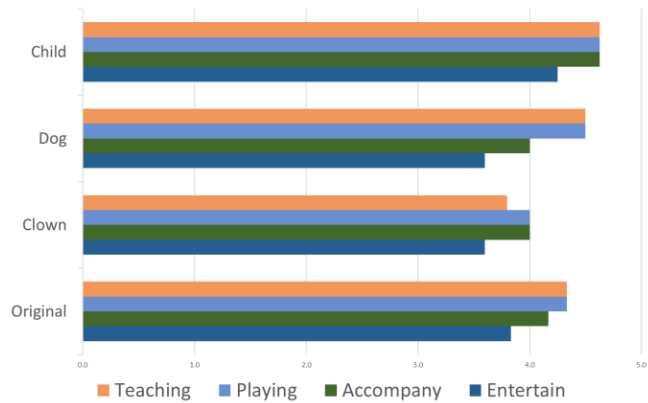

Figure 4: perception of the RMC's role from the online questionnaire.

IV. CONCLUSIONS

This study used mixed methods to gather children's needs, perceptions, and preferences regarding RMCs' functionality, behaviors, and appearance. We found that children and teenagers accept using robots as medical clowns. In addition, certified professional medical clowns were also receptive to this concept. The desire for a human-resemble RMC was found in all four steps of the study and among all user groups. Furthermore, the appearance of the RMC affects the children's perception of its characters and role; the child-like robot was the preferred design and was also perceived as the one who could fulfill different roles. On the opposite, the clown-like robot was perceived as less suitable for different tasks. In addition, our study demonstrated that current robots could be adjusted to be used as RMCs using some design adjustments. Future work should investigate the interaction in hospitals.

V. ACKNOWLEDGMENTS

We would like to acknowledge the contribution of our undergraduate student Saar Weiss for the development of the temi RMC prototype and the execution of the interviews, focus groups, and one-on-one interactions.

This research was supported by the Ministry of Innovation, Science and Technology, Israel (grant 3-15625), and by Ben-Gurion University of the Negev through the Helmsley Charitable Trust, the Agricultural, Biological and Cognitive Robotics Initiative, the W. Gunther Plaut Chair in Manufacturing Engineering and by the George Shrut Chair in Human Performance Management.

Come visit our poster at ICRA 2023. 15:00 - 16:40 | Tue 30 May | PH Stands 1-25 | TuPOS2S-24.12